\documentclass[conference]{IEEEtran}
\IEEEoverridecommandlockouts
\usepackage{amsmath,amssymb,amsfonts}
\usepackage{algorithmic}
\usepackage{graphicx}
\usepackage{textcomp}
\usepackage{xcolor}
\usepackage{pgf}
\usepackage{comment}
\usepackage{nameref}
\usepackage{soul}
\setstcolor{red}
\DeclareMathOperator*{\argmax}{argmax}

\def\BibTeX{{\rm B\kern-.05em{\sc i\kern-.025em b}\kern-.08em
    T\kern-.1667em\lower.7ex\hbox{E}\kern-.125emX}}
\begin{document}

\title{Improving Robustness of Deep Reinforcement Learning Agents: Environment Attack based on the Critic Network}

\author{\IEEEauthorblockN{Lucas SCHOTT}
\IEEEauthorblockA{
\textit{IRT SystemX, ISIR, Sorbonne University}\\
Palaiseau, France \\
lucas.schott@irt-systemx.fr}
\and
\IEEEauthorblockN{Hatem HAJRI}
\IEEEauthorblockA{\textit{IRT SystemX}\\
Palaiseau, France \\
hatem.hajri@irt-systemx.fr}
\and
\IEEEauthorblockN{Sylvain LAMPRIER}
\IEEEauthorblockA{\textit{ISIR, Sorbonne University}\\
Paris, France \\
sylvain.lamprier@isir.upmc.fr}
}

\maketitle

\begin{abstract}
To improve robustness of deep reinforcement learning agents, a line of recent works focus on producing disturbances of the dynamics of the environment. Existing approaches of the literature to generate such disturbances are environment adversarial reinforcement learning methods. These methods set the problem as a two-player game between the protagonist agent, which learns to perform a task in an environment, and the adversary agent, which learns to disturb the dynamics of the considered environment to make the protagonist agent fail. Alternatively, we propose to build on gradient-based adversarial attacks, usually used for classification tasks for instance, that we apply on the critic network of the protagonist to identify efficient disturbances of the dynamics of the environment. Rather than training an adversary agent, which usually reveals as very complex and unstable, we leverage the knowledge of the critic network of the protagonist, to dynamically increase the complexity of the task at each step of the learning process. We show that our method, while being faster and lighter, leads to significantly better improvements in robustness of the policy than existing methods of the literature.
\end{abstract}


\section{Introduction}

    \label{intro}
    
    \footnotetext{Published in the International Joint Conference on Neural Networks (IJCNN 2022)}

    Reinforcement Learning (RL) is a set of methods widely used to train autonomous agents in many different contexts. Autonomous agents trained with these methods can be very effective and reach super-human performances at the task they have been trained for \cite{mnih2013playing}. However, when the environment conditions change, even slightly, the agent performances can drop dramatically \cite{pinto2017robust}. These issues can be observed, for instance, when going from the simulation to the real world \cite{ma2018improved}. One of the main challenges in autonomous systems is to obtain policies that are robust to disturbances and changes in the environment, in particular to cope with reality gap issues.
    
    To improve robustness of autonomous agent policies, a line of approaches focus on Adversarial Reinforcement Learning methods \cite{pinto2017robust,heinrich2015fictitious,heinrich2016deep,ma2018improved}. In particular on methods formulated as two-player games between a protagonist agent, which learns to perform a task, and an adversary agent, which learns to generate efficient disturbances in the environment to make the protagonist agent fail. The adversary can for instance: generate disturbances of the dynamics of the environment, move obstacles, or change the topology of the environment. The goal is to train the protagonist agent to become more robust to the disturbances generated by the adversary, assuming that this can generalize in slightly different, possibly more complex, environments in which it can be deployed. In this paper, we propose a new approach which is to rely on Adversarial Attacks of Neural Networks, and in particular based on the Critic Network of the protagonist agent, to generate meaningful disturbances of the dynamics of the environment, without having to train an adversary agent.
    
    Our contribution is :
    \begin{itemize}
        \item Applying adversarial attacks of neural networks to generate disturbances of the dynamics of the environment.
        \item Using the gradient of the critic Network to generate meaningful long term based disturbances.
        \item Showing the efficiency of our approach compared to existing methods that have to train an adversary agent to generate disturbances of the environment.
        \item Showing the benefits of attacking the environment rather than attacking the observations to improve the robustness of the policy.
    \end{itemize}

\section{Backgrounds and Task}

    \subsection{Reinforcement Learning}
    
        Reinforcement learning \cite{sutton1998introduction} consists of an agent which sequentially chooses actions from observations over a sequence of time steps and learns to perform a task by trying to maximize cumulative discounted rewards.
        
        The problem is modeled as a Markovian Decision Process (MDP) \cite{bellman1957markovian} which can be expressed by the tuple $(S,A,P,R)$,  where $S$ is the state space, $A$ is the action space, $P : S \times A \times S \rightarrow [0;1]$ is the transition probability $P(s_{t+1}|s_t,a_t)$. $R : S \times A \times S \rightarrow {\rm I\!R}$ is the reward function $R(s_t,a_t,s_{t+1})$. In general, agents only get a partial observation $x_t = X(s_t)$ of the state $s_t\in S$ at time step $t$, $X$ being the observation function. In this paper we assume that the training environment is fully-observed  (i.e., $\forall (s,s'), X(s)=X(s') \Rightarrow s=s' $). In the following we will use $x$ to denote the inputs of the neural networks of the agent.
        
        RL algorithms attempt to learn a policy $\pi_\theta : X \times A \rightarrow [0;1]$ which calculates the probability  $\pi_\theta(a_t|x_t)$ to select the action $a_t$ given the observation $x_t$, $\theta$ denotes the parameters for the policy $\pi$. The goal is to maximize the expected cumulative discounted reward $\mathbb{E}_{\tau \sim \pi_\theta}  \sum^{\infty}_{t=1} \gamma^{t}r_t$ with $r_t$ the reward obtained at step $t$ in episode $\tau$, $\gamma \in ]0,1[$ being the discount factor. The policy $\pi_\theta$ is learned from sequential experiences in the form of transition tuples $(x_t,a_t,r_{t+1},x_{t+1})$.
        
        In this paper we work with the Proximal Policy Optimization algorithm (PPO) \cite{schulman2017proximal}, because it is composed of a critic network $V_\phi$ in addition to the actor network $\Pi_\theta$. The critic network outputs an estimation of the value function $V_\phi(x_t)$ of any observation $x_t$. This corresponds to the expected cumulative future discounted rewards starting from the state $s_t$ represented by the observation $x_t$, following the current policy $\pi_\theta$ defined by the actor network. In our main contribution, we rely on the knowledge carried by this critic network to build our attacks. Any other RL approach could be considered, provided that a critic network is added and trained on a replay from experiences of the agent following its own policy $\pi_\theta$.

    \subsection{Task: Anticipate Environment Evolution}
    \label{task}
    
    To learn robust policies with reinforcement learning, several approaches have been explored. The robust algorithm approach is to use an algorithm which, due to its structure, will learn a policy by minimizing the risks. An example is the RS-DQN method \cite{fischer2019online}. Other approaches are robust training methods that aim to robustify the policy of an agent by interfering with the rewards during training. Among them cite for instances methods like RLPR \cite{wang2020reinforcement} and AMCS \cite{huang2019deceptive}. These approaches, enable the agent to learn safe policy if the environment stay unchanged, but they are not designed to learn a policy which would be robust to changes in the environment.
    
    As mentioned in \nameref{intro} (section \ref{intro}), an important problem when deploying in real-world an RL agent trained in simulation, is that the gap between the corresponding MDPs can make the policy very ineffective.

    More formally, consider that the target environment is an MDP $\Omega^*=(S^*,A,P^*,R^*)$, and assume that the only environment we have access during learning is defined by the MDP $\Omega=(S,A,P,R)$. We assume that the simulator is able to represent every target state (i.e., $S^* \subset S$), but some of the states in $S^*$ are not reachable following $P$ in $\Omega$ given the  distribution of initial states. Our aim is to learn an agent from $\Omega$ which is robust to different conditions from $\Omega^*$. We assume that we have no knowledge about the  changes that will be applied on $\Omega$, but that $P^*$ follows  dynamics similar to those of  $P$ (e.g., same physical equations). We also assume that $S$ is too large to consider a naive solution which would consist in uniformly sampling from $S$ to learn a robust agent (which would be particularly inefficient, with many unrealistic situations). Thus, to anticipate possible evolution of the environment conditions, we need to consider slight changes around situations reached during learning. This is the purpose of adversarial reinforcement learning approaches, that we focus on in this paper. In section \ref{contrib} \nameref{contrib}, we assume that we have access to features of the simulator that allow us to manipulate states $s \in S$, by applying slight modifications to obtain new states $s'$ respecting  constraints from the environment (i.e., $s' \in S$).

\section{Related Work}    
    
    Adversarial Reinforcement Learning methods seek at improving robustness of neural network policies to various type of threats: like corrupted inputs, unknown situations, uncertainty in the environment, increasing difficulty of the environment. These type of threats have been addressed with two different type of approaches :
    
    The first approach, to address corrupted inputs, is to generate perturbations in the observation space of the agent \cite{behzadan2017vulnerability}, i.e. produce a adversarial observation $x'$ from an original observation $x$ in order to fool the agent. With the objective to make the protagonist more robust w.r.t corrupted observations by training it on adversarial examples $x'$. For example, in autonomous systems, this may concern corruptions due to sensor failures.
    
    The second approach, to address unknown situations, uncertainty in the environment and increasing difficulty of the environment, is to generate disturbances of the dynamics of the environment \cite{pinto2017robust,ma2018improved}. While the first approach which generates perturbations in the observations only changes any $x$ to a given $x'$ with no direct implications in the environment, generating disturbances in the environment changes any current state from $s$ to $s'$, changing the dynamics of the environment and leading the agent to a new situation to solve. A given disturbed state $s'$ remains a valid state of the environment, but may be a state impossible to reach following its original dynamics and initial conditions. Since the set of such possible modifications may be very large, there is a need for strategies to generate great disturbances leading to useful disturbed states $s'$, which corresponds to difficult situations for the agent.
    
    In the next section \ref{RARL_FSP}, we present state of the art Environment Adversarial Reinforcement Learning methods that generate disturbances of the dynamics of the environment. And in  section \ref{ObsAttack}, we present Observation Adversarial Reinforcement Learning methods that generates adversarial observations to be provided to the agent.
    
    \subsection{Environment Adversarial Reinforcement Learning}
        \label{RARL_FSP}
        
        Environment Adversarial Reinforcement Learning methods seek at improving robustness of neural network policies by formulating the learning problem as a two-player game. While the protagonist agent learns to solve the task, an adversary agent learns to generate disturbances of the dynamics of the environment to perturb the protagonist.
            
        \subsubsection{Robust Adversarial Reinforcement Learning}
            (RARL) \cite{pinto2017robust} is modeled as a two-player zero-sum Markov Game \cite{littman1994markov} in which the protagonist and the adversary are co-trained by reinforcement learning. The MDP of this game is expressed as the tuple $(S,A_p,A_a,P,R_p)$, where $S$ is the state space, $A_a$ and $A_p$ are the action spaces of the adversary and the protagonist. $P : S \times A_a \times A_p \times S \rightarrow [0;1]$ is the transition probability and $R_p : S \times A_a \times A_p \times S \rightarrow {\rm I\!R}$ is the reward function of the protagonist agent. In a zero-sum two-player game, the reward received by the protagonist is $R_p$, while the adversary gets the reward $R_a = -R_p$. \texttt{RARL} can also be expressed in a semi-cooperative setting \cite{ma2018improved}, the adversary getting partial cooperative rewards to generate more realistic disturbances. In the semi-cooperative setting the reward of the adversary is $R_a =  \alpha R_c - (1-\alpha) R_p$, where $R_c$ is a cooperative reward to reduce the amplitude of the disturbances, and $\alpha$ is the cooperation coefficient, which is used to regulate the proportion of cooperative rewards compared to competitive rewards.

        \subsubsection{Fictitious Self Play}
            (FSP) \cite{heinrich2015fictitious,heinrich2016deep} follow the same principle as \texttt{RARL}, except that the adversary is not just an RL agent, but is an agent which uses two models. The adversary agent uses an RL Model which learns to produce the most efficient disturbances against the protagonist agent, and a Supervised Learning Model which learns to produce the average historical strategy of the RL model. During the training time of the protagonist agent, the adversary produces attacks based on its RL model and attacks based on its SL model. The goal of this method is to prevent the protagonist agent from over-fitting its defense strategy on the best strategy of the adversary agent, and then to better generalize to other type of disturbances and keep good performances in situations without disturbances. \texttt{FSP} can also be expressed in a semi-cooperative setting \cite{ma2018improved}.
        
            \cite{ma2018improved} studied and compared \texttt{RARL} and \texttt{FSP}, they have empirically showed that \texttt{FSP} with semi-cooperative setting is the best method to train an agent to be robust to changes in the environment. \texttt{FSP} with its stochasticity leads to a better adversarial training than \texttt{RARL}, without over-fitting on generated disturbances. And the semi-cooperative setting enables the disturbances to be more realistic and leads to better anticipations of possible changes in the environment. In section \ref{results} we implement and test these methods to see if we can reproduce the results of \cite{ma2018improved} and compare them with our own method.

    \subsection{Observation Adversarial Reinforcement Learning}
        \label{ObsAttack}

        Observation adversarial reinforcement learning approaches, which are related to our work, relies on gradient-based adversarial attacks, classically applied on Neural Networks Classifiers (NNC). Gradient-based adversarial attacks of NNC arise from \cite{goodfellow2014explaining}, which showed that NNC are vulnerable to adversarial inputs. Small perturbations of an images, even imperceptible for human sight can fool an NNC and make it misclassify the adversarial example. Several methods have been proposed, various adversarial attacks of NNC depending on different metrics $L_0$, $L_2$ or $L_{\infty}$, see for instance the survey \cite{2018arXiv181000069C}.
        
        Applying adversarial attacks on Neural Networks Policies (NNP) of RL agents consists of crafting an adversarial observation $x'$ to replace the original observation $x$ for the agent. Then the agent choose the action based on the adversarial observation $a' = \pi(x')$, rather than on the original observation $a = \pi(x)$. For an agent with a discrete action space, an attack is successful if the agent changes its decision $a' \neq a$. By having its observations attacked, the agent is led to take bad decisions, and then it can fail to achieve its goal. Any adversarial attack method developed to fool NNC is usable to fool neural network policies trained by reinforcement learning with discrete action space. It has been shown that NNP are vulnerable to adversarial attacks which perturb the observations of the agent in a RL setting \cite{huang2017adversarial}. Introducing adversarial observations during the training of an RL agent helps the agent to become more robust to corrupted inputs \cite{behzadan2017vulnerability}.
        
        We notice that, to the best of our knowledge, adversarial attacks of neural networks have only been used to produce adversarial observations, but have never been used to generate disturbances of the dynamics of the environment, which we propose to consider in the next section \ref{contrib}.

\section{Contributions}
    \label{contrib}

    \subsection{Adversarial Attacks of Neural Network Applied to the Dynamics of the Environment}
        \label{attack_env}
        
        We propose a new approach which uses adversarial attacks of neural networks to disturb the dynamics of the environment by modifying the states $s$ of the environment. The goal is to perform an adversarial training of the protagonist agent against focused disturbances in the environment, hence improving the robustness of the agent's policy to changes in the environment.
        
\begin{figure}[!ht]
            \centering
            \includegraphics[width=\columnwidth]{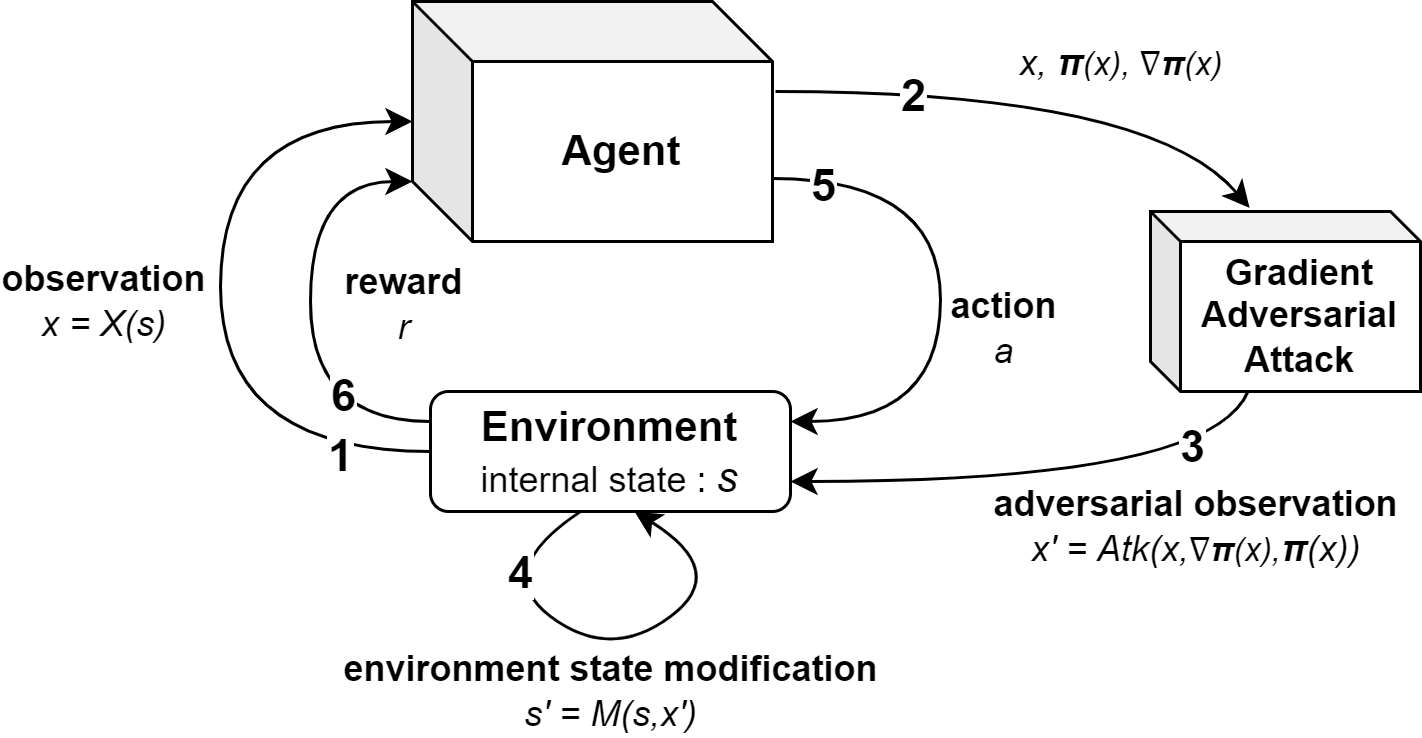}
            \caption{Scheme of the adversarial training of an agent against adversarial attacks of neural networks applied to dynamics of the environment.}
            \label{fig:env_adv_attack}
        \end{figure}

        The principle of our approach is: to use a gradient based adversarial attack of neural network $Atk$ which generates adversarial observations $x' = Atk(x,\pi(x))$, and to use an environment modifier function $M$ which uses the adversarial observation $x'$ and the current state $s$ to modify the environment in a manner that the state of the environment $s$ actually changes to a state $s' = M(s,x')$, with $X(s') \simeq x'$. This means that our proposed method produces a disturbed state $s'$ which is a state of the environment which could be the source of the adversarial observation $x'$. The goal is to disturb the dynamics of the environment by setting the environment to a disturbed state $s'$ rather than the original state $s$. The function $M$ which does the modification of the environment shall modify the environment in a realistic manner, such as the disturbed state $s'$ is still a valid state of the environment $s' \in S$. The scheme of the adversarial training process we propose, as just described, is shown in Fig. \ref{fig:env_adv_attack}.
        
        The method we propose thus allows to use any existing adversarial attack methods of neural networks which generate adversarial observations $x'$. And it produces disturbances of the dynamics of the environment by setting the state of the environment to $s'$ with $X(s') \simeq x'$. To do so, we need to focus on environments with observation spaces which are mapped to features spaces of the environment that one can manipulate from disturbances crafted from observations. This thus requires to know the mapping function from observations to features from the environment. For simplicity purposes, we discuss about environments with direct one to one connections between observations and features of the environment. But this could be extended to more complex environment modifier functions to include in $M$. Please note that this knowledge of the mapping in $M$ is only required during training (i.e., on the simulator), not anymore when the agent is deployed in the target environment, since only used on the inputs of the critic network (to disturb the simulator environment) and not on those of the actor policy.

    \subsection{Adversarial Attacks on Actor-Critic Agents}
    \label{EnvAttack}

        As explained in section \ref{ObsAttack}, adversarial attack methods on NNP are used to change agents decisions by perturbing its inputs $x$. As these methods were designed to fool classifiers \cite{goodfellow2014explaining,2018arXiv181000069C}, they could directly be applied in the context of RL for disturbing observations $x$ given to the agent  \cite{huang2017adversarial,behzadan2017vulnerability}. However, as mentioned in section \ref{attack_env}, in the setting of this paper our goal is rather to create meaningful disturbance of the dynamics of the environment, in order to generate difficult situations $s'$ that the agent has to cope with. In section \ref{EAAN}, we first discuss of an approach based on the actor network of the protagonist agent, which presents limitations, that we overcome in section \ref{EACN} with the second approach based on the critic network of the agent, that corresponds our main contribution.
        
            \subsubsection{Environment Attack based on the Actor Network (EAAN)}
                \label{EAAN}
        
                The first method we could consider to generate adversarial examples is to attack the actor network of the agent. As done in works crafting attacks on observations given an actor policy \cite{huang2017adversarial}, we seek at crafting an adversarial observation $x'$, by applying a perturbation $\eta$ on $x$, that minimizes the probability of the dominant action $a_d$. The principle of our method, inspired by \cite{papernot2016limitations} (see also \cite{cesaire2021stochastic} for a recent improvement), is to leverage the knowledge of the Jacobian matrix of the function learned by the actor network with respect to the inputs to generate perturbations. Consider the actor network $\Pi_\theta$, and denote by $\Pi(x)$ its logit outputs, $\Pi_{a_j}(x)$ is the probability of the action $a_j$ given the input $x$. To craft an adversarial example $x'$ from a given input $x$ at a distance of $||x'-x|| = ||\eta|| = \varepsilon$, we first define a perturbation map $H$, based on the jacobian $\nabla_x \Pi(x)$, as: 
                \begin{equation}
                    \label{eq:saliency-map-decreasing-features}
                    H[i] = \left( \sum_{\substack{a_j\neq a_d \\ a_j\in A}} \frac{\partial \Pi_{a_j}(x)}{\partial x_i}\right) - \left(  \frac{\partial \Pi_{a_d}(x)}{\partial x_i}\right)
                \end{equation}
                where $a_d = \argmax\limits_{a \in A} \Pi_a(x)$ is the dominant action for $\pi_\theta$ given the observation $x$. Assuming local smoothness of the policy function, this enables to generate the normalized perturbation $\eta$ as:
                \begin{equation}
                    \eta = \frac{\varepsilon}{||H||} H
                \end{equation}
                Then, $\eta$ is used to create the adversarial observation $x'$:
                \begin{equation}
                    x' = x + \eta
                \end{equation}
                This method creates an adversarial example $x'$ which is an observation which reduces the probability of the dominant action $a_d$ and increases the probability of all other actions. Finally, the idea is to modify the environment to set the current state to $s' = M(s,x')$, with $x' \simeq X(s')$. $s'$ is thus a state in which the agent is less confident in the dominant action for the benefit of the other actions. During the adversarial learning procedure, this lean the sampling process toward situations with greater uncertainty of the agent, possibly outside the original space of reachable states $S^* \subset S$.
                However, attacking the actor network does not allow to choose the state modification with a long term disturbance feedback. It only punctually leads the agent to states where the action taken is non optimal. But these states may be good and safe states anyway, where any action is acceptable. An attack that would consider the expected cumulative reward provided by the critic network would be more effective.
                
            \subsubsection{Environment Attack based on the Critic Network (EACN)}
                \label{EACN}

                In consequence, we propose to rather produce the attack on the base of the critic network (an estimate of the value function) used for example in the PPO algorithm. Given an observation $x_t$, the output of this critic network is a single scalar $V(x_t)$, which is an approximation of the discounted sum of the future expected rewards starting from this observation, $V(x_t) = \mathbb{E} [ \sum^{\infty}_{t=0}{\gamma^{t}r(s_t,a_t)} ]$.
                The method we propose leverage the knowledge of the Jacobian matrix of the value function learned  by the critic network with respect to inputs to generate perturbations. The goal of this attack is to decrease the value of the output of the critic network.
                Consider the critic network and denote by $V(x)$ its output. To craft an adversarial example $x'$ from a given input $x$ with a perturbation $\eta$ at a distance of $||x'-x|| = ||\eta|| = \varepsilon$ , our method first computes the jacobian $\nabla_x V(x)$. Assuming local smoothness of the value function, this enables to generate the normalized perturbation $\eta$ as:
                \begin{equation}
                    \eta = - \frac{\varepsilon}{||\nabla_x V(x)||} \nabla_x V(x)
                \end{equation}
                Then, to decrease the value of the output of the critic network, we generate the adversarial example $x'$ as :
                \begin{equation}
                    x' = x + \eta
                \end{equation}
                This method creates an adversarial example $x'$ for which $V(x') < V(x)$, thus being likely to allow for a lower expected cumulative reward given by the critic, according to the current policy. Finally, the idea is to modify the environment to set the current state to $s' = M(s,x')$. As $X(s') \simeq x'$, $s'$ is thus a state with lower value for the agent. During the adversarial learning procedure, this lean the sampling process toward unwanted and difficult situations, possibly outside the original space of reachable states $S^* \subset S$, with greater uncertainty of the agent.

\section{Materials and Method}

    \subsection{Environments}
            
\begin{figure}[!ht]
            \centering
            \includegraphics[width=0.7\columnwidth]{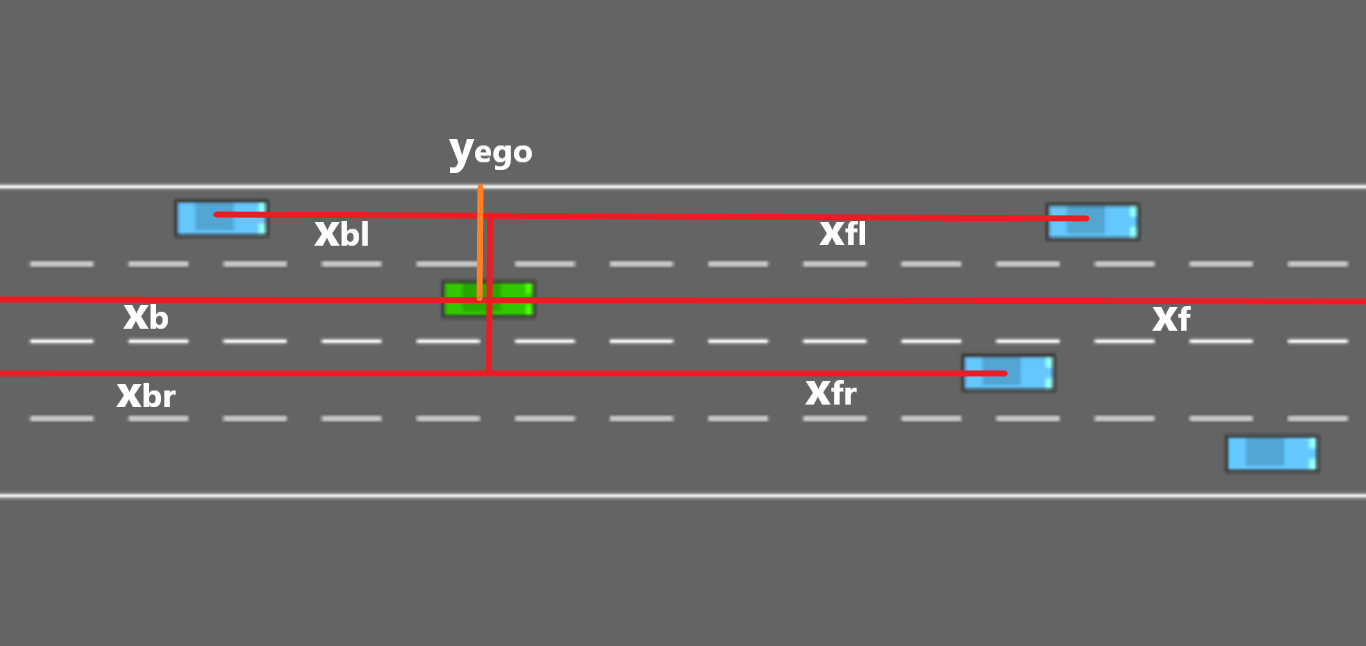}
            \caption{Screen capture and state representation (without velocities) in HighwayEnv}
            \label{fig:capture_highway}
        \end{figure}

        \paragraph{HighwayEnv \cite{leurent_environment_2018}} This is a 2D open-source autonomous driving simulation environment, illustrated in Fig. \ref{fig:capture_highway}, where the agent drives a car on an infinite 4 lanes unidirectional highway. The vehicle piloted by the agent (the ego-vehicle) is inserted in a traffic flow of other vehicles (the exo-vehicles). All exo-vehicles follow a basic driving algorithm. The goal of the agent is to drive as fast as possible without having an accident. The agent observes its own transverse position in the width of the road, its own velocity, and the relative positions and velocities of the exo-vehicles closest to the agent in contiguous lanes (back and front). The action of the agent $a_t \in A$ is a discrete choice between 5 possibilities: (0) change lane to the left; (1) nothing (stay on the same lane, at the same velocity); (2) change lane to the right; (3) accelerate ($+5 m \cdot s^{-1}$); (4) decelerate ($-5 m \cdot s^{-1}$).

        The episode ends when the agent has a collision with another vehicle, or when it reaches the time limit. The goal of the agent is to drive as long as possible without having a collision, and the faster as possible to get the maximum total reward.
            
        The disturbances we apply on the dynamics of the environment HighwayEnv are modifications of the positions and the velocities of the exo-vehicles around the agent. The base configuration of the environment is defined by a low traffic density on the road around the agent, and the environment can be more difficult by increasing the traffic density on the road.
        
\begin{figure}[!ht]
            \centering
            \includegraphics[width=0.35\columnwidth]{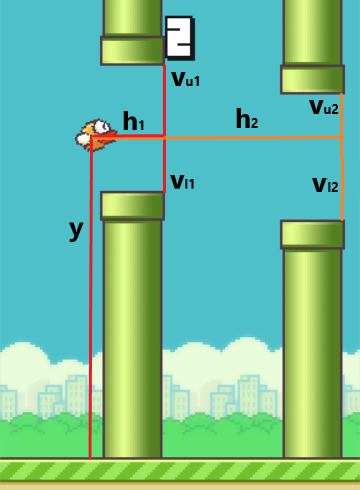}
            \caption{Screen capture and state representation in FlappyBird}
            \label{fig:capture_flappy}
        \end{figure}

        \paragraph{FlappyBird  \cite{nogueira_talendarflappy-bird-gym_2021}} This is a 2D open-source game environment, illustrated in Fig. \ref{fig:capture_flappy}, where the agent is a bird which has to fly as long as possible and avoid obstacles. The agent observes its own position, and the positions the two nearest obstacles in front of him. The action of the agent is a discrete choice between 2 possibilities: (0) do nothing (let the gravity pull down the bird); (1) flap the bird's wings (add an instantaneous upward velocity). The reward function of the agent is at each time step $R = 1$. The episode ends when the bird collides with an obstacle or with the ground, or if the agent reaches the time limit.
        
        The disturbances we apply on the dynamics of the environment FlappyBird are modifications of the vertical position of the obstacles in front of the agent. The base configuration of the environment is defined by large gap size, which is the vertical space between the obstacles, and the environment can be more difficult by decreasing gap size between the obstacles. For the experiments and the evaluations, we follow the methodology as described in the next section \ref{exp}.
        
    \subsection{Methodology}
        \label{exp}
        
        In order to evaluate the robustness of the policy w.r.t. environment changes, we first train our agent in a base configuration of the environment, that represents the simulation environment referred in section \ref{task}. Then, we test our trained agent in the environment, but in another configuration which increases difficulty; this represents the target environment referred in section \ref{task}, which is unknown during training. 
        
        We first pre-train, during 1 million training steps, a protagonist agent in the base environment. This agent is used as pre-trained model for all the other protagonist agents we train. Starting from this pre-trained agent, we train a \texttt{Baseline} agent by continuing training for 2 million steps in the same base configuration. Starting from this pre-trained agent, we also train a \texttt{Target} agent in the complexified target environment.
        
        Also starting from the pre-trained agent, we proceed adversarial training of a protagonist agent for each adversarial attack method, including ours, for 2 million training steps: Environment Attack based on the Critic Network (EACN), Environment Attack based on the Actor Network (EAAN), Observation Attack based on the Critic Network (OACN) and Observation Attack based on the Actor Network (OAAN). While \texttt{EACN} is  the method proposed in this paper, \texttt{EAAN}, \texttt{OACN} and \texttt{OAAN} are ablations, where attacks of performed either on the actor network (rather the critic one), or performed on observations (no direct impact on the dynamics of the environment, as described in section \ref{ObsAttack}), or both.
        
        We also train agents starting from the pre-trained one, for both \texttt{RARL} and \texttt{FSP} methods, described in section \ref{RARL_FSP}. For these methods, we first train during 1 million steps in the base environment an adversary agent to attack the protagonist. Then, during the next 1 million training steps, we train the protagonist agent to defend against the attack of the adversary agent. For completeness, an alternative version of \texttt{FSP}, which we refer as \texttt{CO-FSP}, where the protagonist and the adversary are trained alternatively, 10 000 time steps for the adversary and 10 000 time steps for the protagonist in turn, up to 2 million environment steps, to assess co-evolution of this state-of-the-art approach. 

        All reported results are aggregated over 5 seeds. In all the following graphs present in this article, lines correspond to means over the seeds, the shaded areas correspond to the 95\% confidence intervals calculated with a student test on the seeds, while averaging over 200 episodes for each.

\section{Results : HighwayEnv}
    \label{results}
    
    \subsection{Attacking the agent}
        
\begin{figure}[!ht]
            \centering
            \resizebox{\columnwidth}{!}{\input{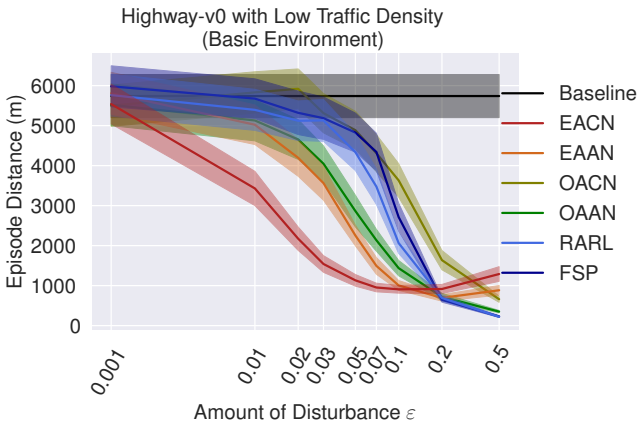}}
            \caption{Performances of the \texttt{Baseline} agent being attacked with the different methods w.r.t. the amount of disturbance $\varepsilon$, in the base configration of HighwayEnv.}
            \label{fig:attack_highway}
        \end{figure}

        First, in figure \ref{fig:attack_highway}, we assess the efficiency of the attacks on the pre-trained undefended agent (whose non-attacked mean reward is referred to as \texttt{Baseline}), according to various amounts of maximal disturbance $\varepsilon$, in the training environment HighwayEnv.
        
        We observe that for low amounts of disturbance $\varepsilon$, our environment adversarial attack methods, \texttt{EACN} and \texttt{EAAN}, are more effective to reduce the performances of the agent than the methods with adversary agents, \texttt{RARL} and \texttt{FSP}. This can be explained by the fact that \texttt{RARL} and \texttt{FSP} need to train an adversary, which corresponds to a hard task for low $\varepsilon$ due to weak possible gains compared to noise from the environment. \texttt{EACN} and \texttt{EAAN} do not suffer from that. On the contrary, for high values of $\varepsilon$, \texttt{RARL} and \texttt{FSP} learn beneficial long-term adversarial policies and generate more effective disturbances in the environment than \texttt{EACN} and \texttt{EAAN}. But high values of $\varepsilon$ greatly increase complexity of the environment with much less realistic environment disturbances. The observation attack methods \texttt{OACN} and \texttt{OAAN} are less effective than \texttt{EACN} and \texttt{EAAN}, because they only cause indirect modifications of future state distributions, by impacting the behavior of the agent. \texttt{OAAN} is more effective than \texttt{OACN}, this shows that attacking the actor network is more effective than attacking the critic network when perturbing the observations. Finally, \texttt{EACN} is more effective than \texttt{EAAN}, this shows that attacking the critic network is more effective than attacking the actor network when disturbing the environment.

    \subsection{Performances of adversarially trained agents}

        \begin{figure}[!ht]
            \centering
            \resizebox{\columnwidth}{!}{\input{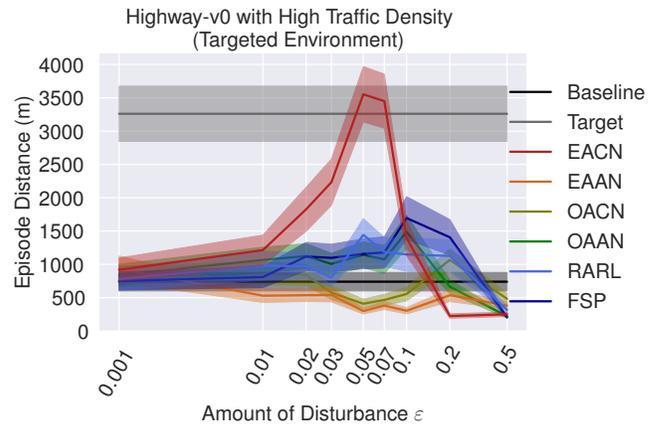}}
            \caption{Performances of the agents, evaluated in the target configuration of HighwayEnv, trained by adversarial training in the base configuration of HighwayEnv against the different attack methods w.r.t the amount of disturbance $\varepsilon$ used during the adversarial training.}
            \label{fig:complex_env_highway}
        \end{figure}
        
        In Fig. \ref{fig:complex_env_highway}, we consider performances in the target environment, of agents adversarially trained in the base environment, according to various amounts of disturbance allowed to attackers. First, we observe that the method we propose, \texttt{EACN}, gets the best results, with a significant and great improvement of the performances of the agent in the target environment for $\varepsilon = 0.05$, reaching and even exceeding the \texttt{Target} agent (trained on the unavailable target environment). \texttt{EACN} produces states that are bad according to the estimation of the long term feedback given by the critic network. States which can either place the agent in immediately difficult situations, or situations that will tend to be difficult to manage in future steps in the training phase. As for other methods, its performances are greatly dependent on the intensity of the attacks, which needs to be carefully tuned via grid-search. However, grid-search for the agents trained with \texttt{EACN}, \texttt{EAAN}, \texttt{OACN} or \texttt{OAAN} are greatly less costly than for \texttt{RARL} or \texttt{FSP}, because \texttt{RARL} and \texttt{FSP} require also to retrain the adversary agent for each item of the grid, in addition to retrain each time the protagonist agent. We notice that \texttt{OAAN} also improves a little the performances of the agent, for $\varepsilon = 0.1$. \texttt{OAAN} creates adversarial observations which are optimized to change the output of the actor network, which leads the agent to take bad decisions and then indirectly reach difficult situations during the training phase. The methods \texttt{EAAN} and \texttt{OACN} do not improve the performances of the agent in the target environment. \texttt{OACN} is ineffective for this purpose because it creates observations that appear dangerous to the agent, while the situation is not in the underlying environment (since the state did not change, only the observations did). Then, the agent only learns to associate observations for dangerous states as safe states, which is counterproductive. And \texttt{EAAN} is ineffective because it produces disturbances which do not create difficult situations, but only unexpected ones, which is not enough to robustify the agent. Finally, we observe that \texttt{FSP} improves more the performances of the protagonist agent than \texttt{RARL}, which is consistent with  results of \cite{ma2018improved}.
        
        We also adversarially train agents with the \texttt{FSP} method, using the best amount of disturbance $\varepsilon = 0.1$, in a semi-cooperative setting. We do an hyper-parameter tuning of the cooperation coefficient $\alpha$, described in \ref{RARL_FSP}, for $\alpha$ from $0$ to $0.5$. We get the best results for $\alpha = 0.2$ where the performances of the agent reach $2000$ meters in the target environment, while with $\alpha = 0$, as in the strictly competitive setting first tested, the agent only reaches $1600$ meters in the target environment. So using semi-cooperative setting increases the performances of the \texttt{FSP} method, which is consistent with \cite{ma2018improved}, but remains significantly less effective than the proposed \texttt{EACN} method.

    \subsection{Training efficiency}
            
\begin{figure}[!ht]
            \centering
            \resizebox{\columnwidth}{!}{\input{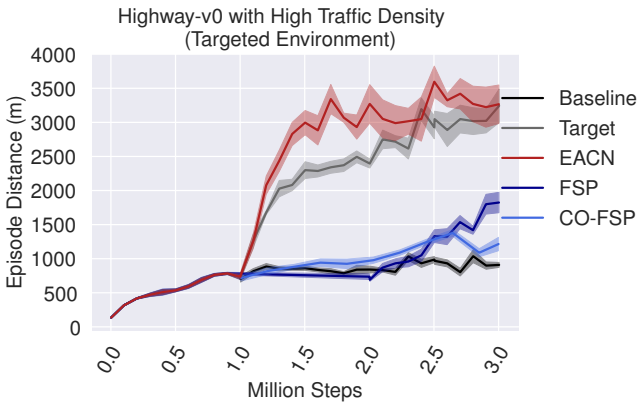}}
            \caption{Evolution of the test performances in the target environment, during adversarial training with the different approaches in the base environment, in HighwayEnv.}
            \label{fig:agent_curve} 
        \end{figure}

        In Fig. \ref{fig:agent_curve}, we report test performance curves of approaches in the target environment, during adversarial training in the base environment, focusing on best parameters for each method. 
        
         We see that after the pre-training phase where no attack is performed (from 0 to 1 million environment steps), our method \texttt{EACN} improves drastically the performances of the agents, compared to the \texttt{Baseline}, which obtain far lower results in the target environment. Both versions of \texttt{FSP} only lead to slight improvements of the performances of the agent, within a greatly longer training horizon than \texttt{EACN}. The curve of \texttt{FSP} is flat from 1 million to 2 million environment steps. This is due to the fact that the performances of the protagonist do not increase during the 1 million training steps of the adversary. After 2 million steps, the protagonist starts the adversarial training against the attacks generated by the adversary and then increases in performances. The performances of the \texttt{CO-FSP} protagonist agent increases directly after 1 million steps, because the adversary and the protagonist are trained alternatively each 10 000 steps. This version remains  however weaker that the former \texttt{FSP} one, which learns to defend against more effective adversary. We also observe that our method \texttt{EACN} enables the trained agents to converge faster than the \texttt{Target} agents, while cannot leverage from training from this unavailable environment. Our hypothesis is that \texttt{EACN} creates a curriculum of learning, that progressively produces more and more difficult situations by manipulating states of the agent, while the reference \texttt{Target} agent has to directly deal with hard situations.

    \subsection{Robustness}

\begin{figure}[!ht]
            \centering
            \resizebox{\columnwidth}{!}{\input{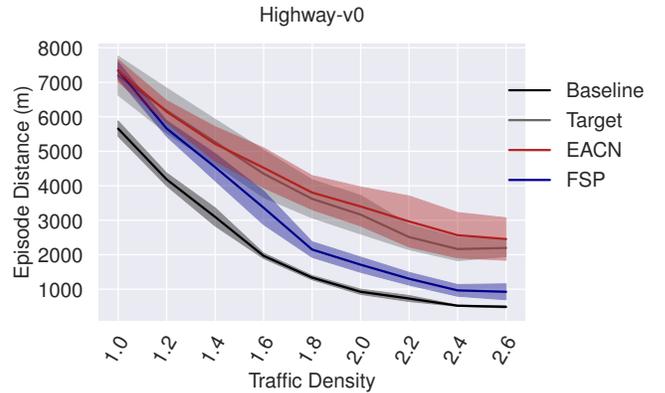}}
            \caption{Performances, w.r.t environment complexity in HighwayEnv.}
            \label{fig:best_agents_highway} 
        \end{figure}

        We finally compare the performances of the agents for varying difficulty settings in the environment HighwayEnv, by changing the traffic density on the road around the agents (Fig. \ref{fig:best_agents_highway}).
        
        We observe that for a low traffic density of $1.0$ (which corresponds to the base environment), agents trained with \texttt{EACN} and \texttt{FSP} methods present similar performances, significantly better than the \texttt{Baseline} agent. While the traffic density increases, the difference between \texttt{EACN} and \texttt{FSP} grows, with finally the agents trained with \texttt{EACN} being significantly better than agents trained with \texttt{FSP} for the high traffic densities. For all levels of traffic density, agents trained with \texttt{EACN} have similar performances as the \texttt{Target} agents. A traffic density of $2.0$ corresponds to the target environment.

\section{Results : FlappyBird}

\begin{figure}[!ht]
            \centering
            \includegraphics[width=\columnwidth]{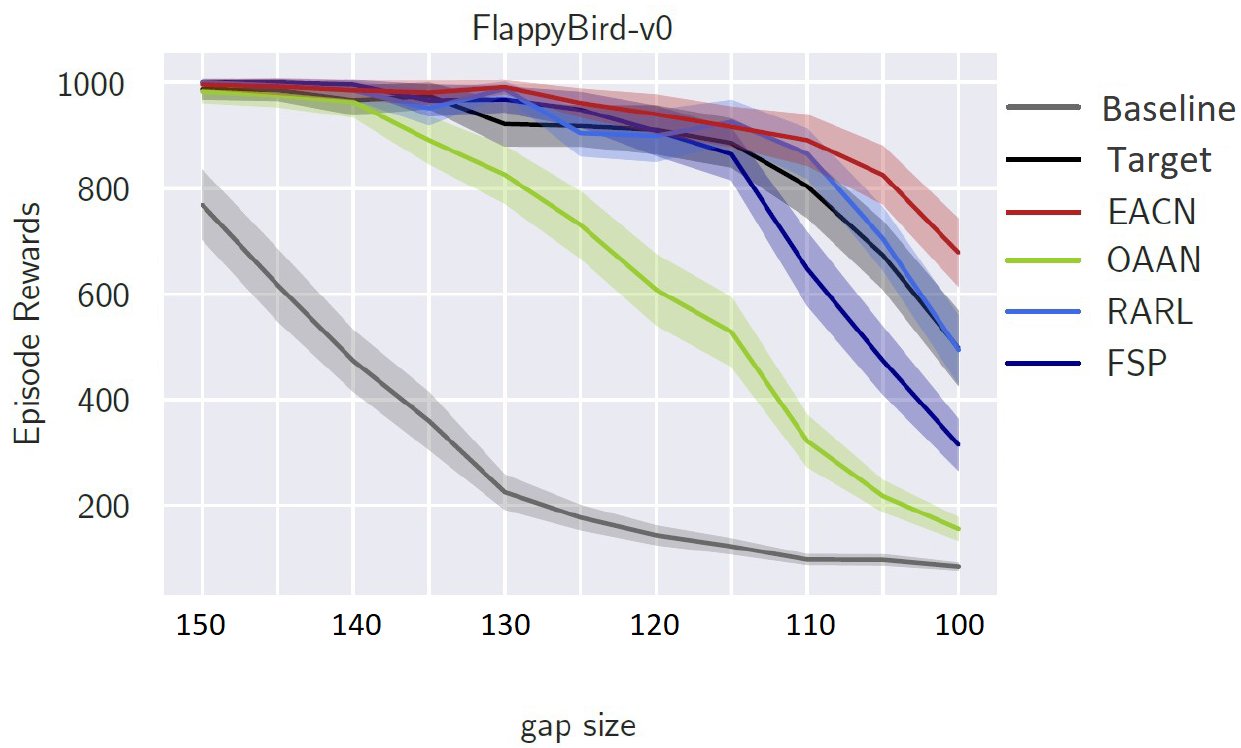}
            \caption{Performances, w.r.t environment complexity in FlappyBird.}
            \label{fig:best_agents_flappybird} 
        \end{figure}

        Here we report final robustness results for the FlappyBird environment, intermediate results being similar to those obtained for HighwayEnv. We report in  Fig. \ref{fig:best_agents_flappybird} the performances of the agents for varying difficulty settings, by decreasing the size of the gap between the obstacles in front of the agent. From the base environment configuration on the left (gap size = 150) to the target one on the right (gap size = 100). We focus on the best methods and on the best parameters for each method, selected in same way as we did for HighwayEnv.
        
        Consistently with HighwayEnv, we observe that for a large gap size between the obstacles (150), agents trained with any adversarial training method present performances similar to the \texttt{Target} agent, significantly better than those obtained by the \texttt{Baseline} one. While the obstacle gap size decreases, which corresponds to an increased difficulty, the difference between \texttt{EACN} and the other methods grows. We finally observe that for small gap sizes (100), agents trained with \texttt{EACN} are significantly better than all the other agents including the \texttt{Target} one.

\section{Conclusion}

    In this paper, we proposed a method for improving robustness of RL agents based on adversarial attacks applied to the dynamics of the environment. In particular, we proposed to craft an adversarial attack based on the critic network, \texttt{EACN}, which gives insights on the long term  impact of candidate attacks, to add disturbances in the environment without having to train an adversary agent as required in greatly more costly  methods such as \texttt{RARL} or \texttt{FSP}. Our experiments demonstrated the efficiency of our approach \texttt{EACN} for learning robust agents, with performances significantly exceeding those of the costly methods \texttt{RARL} and \texttt{FSP} in the considered environment, and even exceeding those obtained by \texttt{Target} agents trained in the target - unavailable - environment. Future works concern the application of the method when the environment features are not directly linked to observations as it is the case in this work, for more complicated functions $M$ that we can learn from experience.

\section{Acknowledgement}

    This work has been supported by EPI project under the supervision of the SystemX Technological Research Institute in collaboration with Apsys, Expleo, Naval Group and Stellantis, and by the French government under the "Investissements d'avenir” program, as part of IRT-SystemX.

\bibliographystyle{plain}
\bibliography{biblio}

\begin{thebibliography}{10}

\bibitem{behzadan2017vulnerability}
Vahid Behzadan and Arslan Munir.
\newblock Vulnerability of deep reinforcement learning to policy induction
  attacks.
\newblock In {\em International Conference on Machine Learning and Data Mining
  in Pattern Recognition}, pages 262--275. Springer, 2017.

\bibitem{bellman1957markovian}
Richard Bellman.
\newblock A markovian decision process.
\newblock {\em Journal of mathematics and mechanics}, 6(5):679--684, 1957.

\bibitem{2018arXiv181000069C}
Anirban {Chakraborty}, Manaar {Alam}, Vishal {Dey}, Anupam {Chattopadhyay}, and
  Debdeep {Mukhopadhyay}.
\newblock {Adversarial Attacks and Defences: A Survey}.
\newblock {\em arXiv e-prints}, page arXiv:1810.00069, September 2018.

\bibitem{cesaire2021stochastic}
Manon Césaire, Lucas Schott, Hatem Hajri, Sylvain Lamprier, and Patrick
  Gallinari.
\newblock Stochastic sparse adversarial attacks.
\newblock In {\em 2021 IEEE 33rd International Conference on Tools with
  Artificial Intelligence (ICTAI)}, pages 1247--1254, 2021.

\bibitem{fischer2019online}
Marc Fischer, Matthew Mirman, Steven Stalder, and Martin Vechev.
\newblock Online robustness training for deep reinforcement learning.
\newblock {\em arXiv preprint arXiv:1911.00887}, 2019.

\bibitem{goodfellow2014explaining}
Ian~J Goodfellow, Jonathon Shlens, and Christian Szegedy.
\newblock Explaining and harnessing adversarial examples.
\newblock {\em arXiv preprint arXiv:1412.6572}, 2014.

\bibitem{heinrich2015fictitious}
Johannes Heinrich, Marc Lanctot, and David Silver.
\newblock Fictitious self-play in extensive-form games.
\newblock In {\em International conference on machine learning}, pages
  805--813. PMLR, 2015.

\bibitem{heinrich2016deep}
Johannes Heinrich and David Silver.
\newblock Deep reinforcement learning from self-play in imperfect-information
  games.
\newblock {\em arXiv preprint arXiv:1603.01121}, 2016.

\bibitem{huang2017adversarial}
Sandy Huang, Nicolas Papernot, Ian Goodfellow, Yan Duan, and Pieter Abbeel.
\newblock Adversarial attacks on neural network policies.
\newblock {\em arXiv preprint arXiv:1702.02284}, 2017.

\bibitem{huang2019deceptive}
Yunhan Huang and Quanyan Zhu.
\newblock Deceptive reinforcement learning under adversarial manipulations on
  cost signals.
\newblock In {\em International Conference on Decision and Game Theory for
  Security}, pages 217--237. Springer, 2019.

\bibitem{leurent_environment_2018}
Edouard Leurent.
\newblock An {Environment} for {Autonomous} {Driving} {Decision}-{Making},
  2018.

\bibitem{littman1994markov}
Michael~L Littman.
\newblock Markov games as a framework for multi-agent reinforcement learning.
\newblock In {\em Machine learning proceedings 1994}, pages 157--163. Elsevier,
  1994.

\bibitem{ma2018improved}
Xiaobai Ma, Katherine Driggs-Campbell, and Mykel~J Kochenderfer.
\newblock Improved robustness and safety for autonomous vehicle control with
  adversarial reinforcement learning.
\newblock In {\em 2018 IEEE Intelligent Vehicles Symposium (IV)}, pages
  1665--1671. IEEE, 2018.

\bibitem{mnih2013playing}
Volodymyr Mnih, Koray Kavukcuoglu, David Silver, Alex Graves, Ioannis
  Antonoglou, Daan Wierstra, and Martin Riedmiller.
\newblock Playing atari with deep reinforcement learning.
\newblock {\em arXiv preprint arXiv:1312.5602}, 2013.

\bibitem{nogueira_talendarflappy-bird-gym_2021}
Gabriel Nogueira.
\newblock flappy-bird-gym, 2021.

\bibitem{papernot2016limitations}
Nicolas Papernot, Patrick McDaniel, Somesh Jha, Matt Fredrikson, Z~Berkay
  Celik, and Ananthram Swami.
\newblock The limitations of deep learning in adversarial settings.
\newblock In {\em 2016 IEEE European symposium on security and privacy
  (EuroS\&P)}, pages 372--387. IEEE, 2016.

\bibitem{pinto2017robust}
Lerrel Pinto, James Davidson, Rahul Sukthankar, and Abhinav Gupta.
\newblock Robust adversarial reinforcement learning.
\newblock In {\em International Conference on Machine Learning}, pages
  2817--2826. PMLR, 2017.

\bibitem{schulman2017proximal}
John Schulman, Filip Wolski, Prafulla Dhariwal, Alec Radford, and Oleg Klimov.
\newblock Proximal policy optimization algorithms.
\newblock {\em arXiv preprint arXiv:1707.06347}, 2017.

\bibitem{sutton1998introduction}
Richard~S Sutton, Andrew~G Barto, et~al.
\newblock {\em Introduction to reinforcement learning}, volume 135.
\newblock MIT press Cambridge, 1998.

\bibitem{wang2020reinforcement}
Jingkang Wang, Yang Liu, and Bo~Li.
\newblock Reinforcement learning with perturbed rewards.
\newblock In {\em Proceedings of the AAAI Conference on Artificial
  Intelligence}, volume~34, pages 6202--6209, 2020.

\end{thebibliography}

\end{document}